\documentclass[preprint,12pt]{elsarticle}

\usepackage{float}
\usepackage{natbib}
\usepackage{graphicx}
\usepackage{subfigure}
\usepackage{epsfig}
\usepackage{multirow}
\usepackage{multicol}
\usepackage{makecell}
\usepackage{array}
\usepackage{diagbox}
\usepackage{amssymb}
\usepackage{color}
\usepackage{algorithm} 
\usepackage{algpseudocode}

\usepackage{mathrsfs}
\usepackage{amsmath}
\usepackage{amssymb}
\usepackage{extarrows}
\usepackage{setspace}

\usepackage{epstopdf}  
\usepackage{amsthm}
\usepackage{epsfig}
\usepackage{dsfont}

\begin{document}

\begin{frontmatter}

\title{Scalable Angular Discriminative Deep Metric Learning\\
for Face Recognition}


\author[a]{Bowen Wu}
\author[b]{Huaming Wu}
\author[a]{Monica M.Y. Zhang \corref{cor}}

\address[a]{Center for Combinatorics, Nankai University, Tianjin 300071, China}
\address[b]{Center for Applied Mathematics, Tianjin University, Tianjin 300072, China}
\cortext[cor]{Corresponding author. \\ \indent \indent E-mail address: wbw@mail.nankai.edu.cn (B. Wu)\\ \indent \indent \indent \indent \indent whming@tju.edu.cn (Huaming Wu)\\ \indent \indent \indent \indent \indent monicazhang@mail.nankai.edu.cn (Monica M.Y. Zhang)}

\begin{abstract}

With the development of deep learning, Deep Metric Learning (DML) has achieved great improvements in face recognition. Specifically, the widely used softmax loss in the training process often bring large intra-class variations, and feature normalization is only exploited in the testing process to compute the pair similarities. To bridge the gap, we impose the intra-class cosine similarity between the features and weight vectors in softmax loss larger than a margin in the training step, and extend it from four aspects. First, we explore the effect of a hard sample mining strategy. To alleviate the human labor of adjusting the margin hyper-parameter, a self-adaptive margin updating strategy is proposed. Then, a normalized version is given to take full advantage of the cosine similarity constraint. Furthermore, we enhance the former constraint to force the intra-class cosine similarity larger than the mean inter-class cosine similarity with a margin in the exponential feature projection space. Extensive experiments on Labeled Face in the Wild (LFW), Youtube Faces (YTF) and IARPA Janus Benchmark A (IJB-A) datasets demonstrate that the proposed methods outperform the mainstream DML methods and approach the state-of-the-art performance.

\end{abstract}


\end{frontmatter}


\section{Introduction}
Face recognition has been one of the most challenging and attractive areas in computer vision, due to its close relationship with some actual applications, such as biometrics and surveillance. However, face recognition problem is far from solved, since it is closely related to face detection, face alignment, feature extraction (or face representation) and classification, which influence the final performance from different aspects. Especially, feature extraction plays a paramount role. Conventional feature extraction methods (such as LBP, Gabor and SIFT) always work with suitable metric distances (such as Euclidean distance and cosine distance). However, these methods are not discriminative enough to meet the demands for more complex face recognition scenarios. And the situation may be worse when accompanied by inappropriate metric distances.

\begin{figure}[h]
\centering
\vspace{-0.3cm}
 \includegraphics[height=8cm,width=12cm]
 {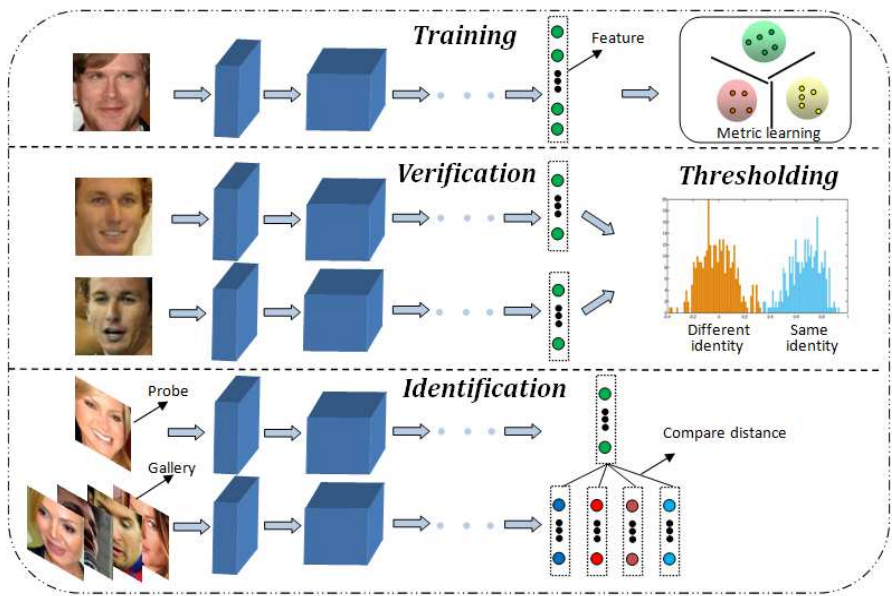}
 \caption{The face recognition pipeline in this paper.}\label{fig1}
\end{figure}

Convolutional Neural Network (CNN), which emerges as a powerful feature extraction method, has drawn much attention due to its excellent performance in computer vision community. Several Deep Metric Learning (DML) methods, which unify deep learning and metric learning into a joint learning framework, have been proposed recently and set new state-of-the-arts in various tasks, such as objection classification \cite{oh2016deep}, image retrieval \cite{zhao2015deep}, person reidentification \cite{yi2014deep}, and so on. Specifically, DML has surpassed the humans'abilities on some benchmark datasets in the field of face recognition \cite{sun2014deep,schroff2015facenet,wen2016discriminative}.

Face recognition can be classified into two tasks, namely face identification and face verification (Fig.~\ref{fig1}). The former aims to classify an input image to a specific identity, while the latter is to determine whether a pair of face images is from the same identity or not. In general end-to-end CNN based face recognition training process, Euclidean distance is used to measure the similarities between features. Whereas, the cosine similarity or normalized inner product is widely used in the testing process. As illustrated in \cite{liu2017sphereface}, Euclidean distance or Euclidean margin-based loss is not always suitable for learning discriminative features, and using normalized features to compute the pair similarities for testing can boost the performance. These properties motivate some works \cite{wang2017normface,chunjie2017cosine} to incorporate the cosine similarity constraint into training stage to keep the consistency with testing. One step further, we force the intra-class cosine similarity larger than a given margin in this paper. Combined with the separability of softmax loss, our original method achieves $0.6\%\sim0.8\%$ accuracy improvement on Labeled Face in the Wild (LFW) dataset and $1\%\sim1.5\%$ accuracy improvement on Youtube Faces (YTF) dataset. Some previous works \cite{sun2014deep1,schroff2015facenet} have clarified the importance of hard sample mining procedure in training CNN, but they haven't exhibited the specific comparative experiment results about whether to use it or not. Here, we compare the effect of cosine similarity constraint on the original training set or the misclassified hard samples of softmax loss. For the diversity of data and the ubiquitously heterogeneous distribution, the global cosine similarity is insufficient to faithfully characterize the true feature distance, as stated in \cite{huang2016local}. A self-adaptive margin updating technique is exploited afterwards, so that the local uniqueness of each identity is considered and the human labor of adjusting the margin is largely saved. To acquire more discriminative features, only imposing the intra-class cosine similarity larger than a margin doesn't seem to be the best choice. So we improve the former constraint to a more powerful case, which enforces the intra-class cosine similarity larger than the mean of the nearest neighboring inter-class cosine similarities in the normalized exponential feature projection space.

In conclusion, our major contributions can be summarized as follows:
1) We first propose a novel metric loss function to directly force the intra-class cosine similarity larger than a fixed margin, so that the training process coincides with the normalized testing criterion. 2) We conduct a contrastive experiment to show the effect of a hard sample mining strategy on the proposed loss function. 3) A self-adaptive margin strategy is incorporated to strengthen the supervision in the updated feature space. 4) To avoid the side effects of infinitely growing norm of features, we further normalize the features and weight vectors of softmax loss to a same value in each mini-batch. 5) A more progressive metric loss function to consider the intra-class and inter-class variations simultaneously is proposed to achieve the discriminative features. Finally, we conduct extensive experiments on three face recognition benchmark datasets, namely LFW \cite{huang2014labeled}, YTF \cite{wolf2011face} and IARPA Janus Benchmark A (IJB-A) \cite{klare2015pushing}, to verify the excellent performance of our approaches.

\section{The Proposed Approaches}

In this section, we reveal the existing phenomenon of large intra-class variations in deeply learned features trained by softmax loss, and propose several novel metric loss functions to alleviate this problem.

\subsection{Recalling Softmax Loss}

From the viewpoint of probability, softmax function aims to convert a vector of real weights to a probability distribution. The original softmax loss is the cross entropy of softmax function, which can be written as
\begin{small}
\begin{equation}
\centering
\mathcal{L_{S}}=-\frac{1}{M}\sum^{M}_{i=1}\log\frac{e^{W_{y_{i}}^{T}x_{i}+b_{y_{i}}}}{\sum_{j=1}^{N}e^{W_{j}^{T}x_{i}+b_{j}}},
\end{equation}
\end{small}where $M$ is the number of training samples, $N$ is the number of classes, $x_{i}$ is the feature of the $i$-th sample, $y_{i}$ is the corresponding class label in range $[1,N]$, $W$ and $b$ are the weight matrix and bias vector of the last inner-product layer before softmax loss, $W_{j}$ is the $j$-th column of $W$ and $b_{j}$ is the corresponding bias term. For simplicity, we omit the bias term in the following experiments, as in \cite{wang2017normface}. Understandingly, if all classes are well-separated, $W_{j}$ will roughly correspond to the mean of features in $j$-th class, and it can also be recognized as the center of $j$-th class in general cases.
\begin{figure}[h]
 \centering
\vspace{-0.3cm}
 \subfigure[]{
 \label{a}
 \includegraphics[height=5cm,width=14cm]{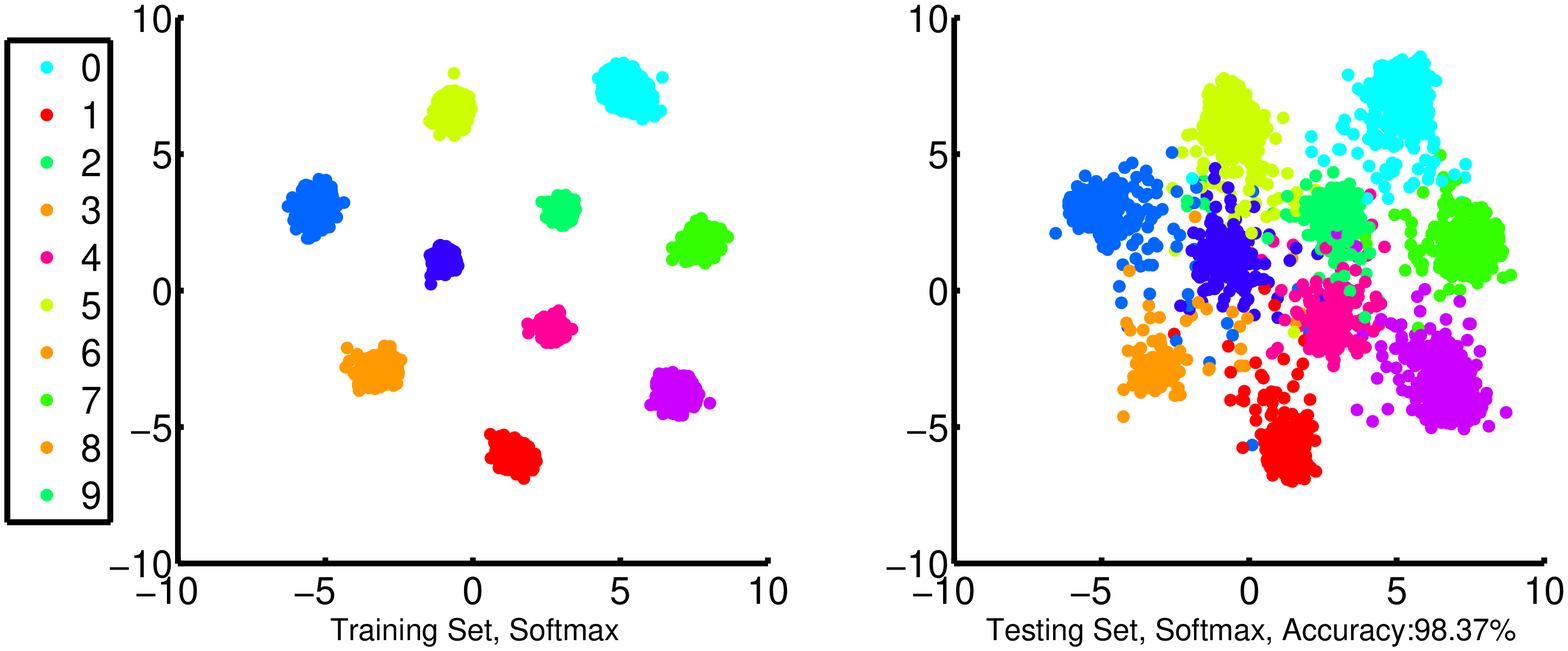}}
 \subfigure[]{
 \label{b}
 \includegraphics[height=5cm,width=14cm]{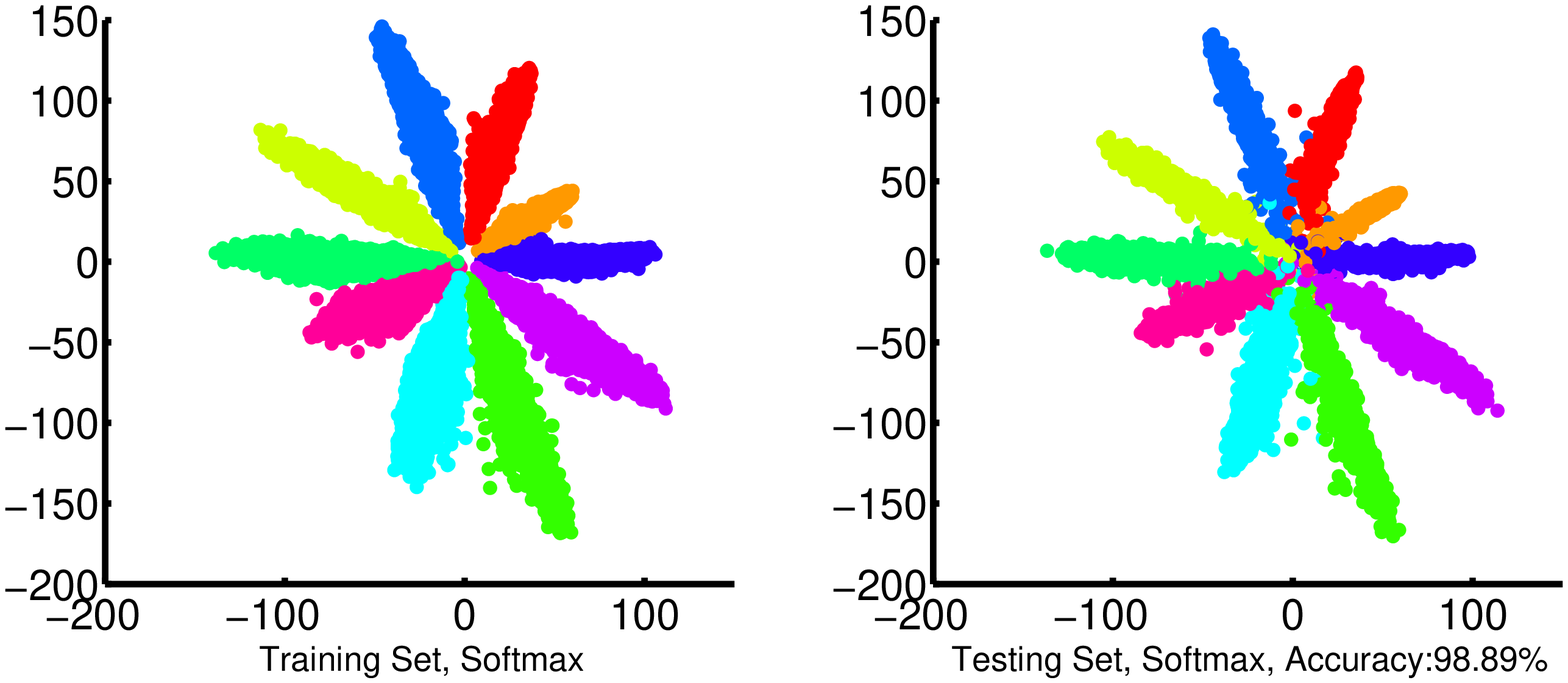}}
 \caption{Visualization of the deeply learned 2-D features on MNIST with (a) MNIST network and (b) LeNet++ network.}\label{fig2}
\end{figure}

To visualize the effect of softmax loss, we conduct a contrastive experiment on the MNIST dataset \cite{lecun1998mnist} with two different CNNs, namely LeNet++ network  \cite{wen2016discriminative} and MNIST network \cite{liu2016large}. We reduce the number of the last feature dimension to 2, so the features can be plotted directly on the 2-D surface. The resulting 2-D features of training and testing sets with the above two different networks are shown in Fig.~\ref{fig2}. We can see that the deeply learned features are separable under the supervision of softmax loss, but not discriminative enough. Especially, there exists significant intra-class variations in the feature space of LeNet++ network, which coincides with the phenomenon elaborated in \cite{wang2017normface} that softmax loss encourages the features to have bigger magnitudes.


\subsection{LMC Loss and HLMC Loss}
To remove the large intra-class variations of softmax loss and to keep the consistency between training and testing, we first propose the Large Margin Cosine (LMC) loss function, which enforces the intra-class cosine similarity between a sample $x_{i}$ and the corresponding weight vector $W_{y_{i}}$ in the last inner-product layer before softmax loss larger than a given margin $\alpha$. The LMC loss function is first formulated as follows:
\begin{small}
\begin{equation}
\centering
\mathcal{L_{C}}=\frac{1}{M}\sum^{M}_{i=1}\left\{\alpha-\widetilde{W}^{T}_{y_{i}}\widetilde{x}_{i}\right\}_{+},
\end{equation}
\end{small}where $\alpha\in [0,1]$, $\widetilde{W}_{y_{i}}=\frac{W_{y_{i}}}{\parallel W_{y_{i}}\parallel_{2}}$ and $\widetilde{x}_{i}=\frac{x_{i}}{\parallel x_{i}\parallel_{2}}$.

Specifically, the joint supervision of softmax loss and LMC loss is necessary to train the CNN for discriminative feature learning. The final LMC loss function for training is
\begin{small}
\begin{equation}
\begin{array}{l}
\mathcal{L_{LMC}}=\mathcal{L_{S}}+\lambda\mathcal{L_{C}}
=-\frac{1}{M}\sum\limits^{M}_{i=1}\log\frac{e^{W_{y_{i}}^{T}x_{i}}}{\sum_{j=1}^{N}e^{W_{j}^{T}x_{i}}}+\frac{\lambda}{M}\sum\limits^{M}_{i=1}\left\{\alpha-\widetilde{W}^{T}_{y_{i}}\widetilde{x}_{i}\right\}_{+},
\end{array}
\end{equation}
\end{small}where $\lambda$ is a weighting parameter that is used for balancing the two loss functions.

It is widely observed that there are often many more easy examples than those meaningful hard ones, an effective data sampling strategy is thus crucial to ensure the learning efficiency of deep features. Therefore, the Hard Large Margin Cosine (HLMC) loss function is explored to impose the previous intra-class cosine similarity constraint on the hard samples. Here, we refer to the hard samples as the ones misclassified by softmax loss, which alleviates the costly computational complexity of pair/triple samples mining strategy adopted in the contrastive/triplet loss \cite{sun2014deep1,schroff2015facenet}.
\begin{small}
\begin{equation}
\begin{array}{l}
\mathcal{L_{HLMC}}=-\frac{1}{M}\sum\limits^{M}_{i=1}\log\frac{e^{W_{y_{i}}^{T}x_{i}}}{\sum_{j=1}^{N}e^{W_{j}^{T}x_{i}}}+\frac{\lambda}{M}\sum\limits^{M}_{i=1}\gamma_{i}\left\{\alpha-\widetilde{W}^{T}_{y_{i}}\widetilde{x}_{i}\right\}_{+},
\end{array}
\end{equation}
\end{small}where $\gamma_{i}\in\{0,1\}$ is a misclassified sample indicator, and $\gamma_{i}=1$ if $x_{i}$ is the misclassified sample of softmax loss.

\subsection{MALMC Loss}

Previous metric loss functions like contrastive loss, triplet loss and L-Softmax loss, often bring in additional hyper-parameters as their fixed margins throughout the training. An intractable hyper-parameter searching process is crucial to the successful training. Following the work \cite{yi2014learning}, we have to suspend training and search for a new margin for every several epochs. In this part, we provide a Margin-Adaptive Large Margin Cosine (MALMC) method, which gives each class an independently updated margin and set it as the maximum of an initially given value and the mean of $p\times Intra(j)$ largest intra-class cosine similarities in the mini-batch.
\begin{small}
\begin{equation}
\begin{array}{l}
\mathcal{L_{MALMC}}=
-\frac{1}{M}\sum\limits^{M}_{i=1}\log\frac{e^{W_{y_{i}}^{T}x_{i}}}{\sum_{j=1}^{N}e^{W_{j}^{T}x_{i}}}+\frac{\lambda}{M}\sum\limits^{M}_{i=1}\left\{\alpha_{y_{i}}-\widetilde{W}^{T}_{y_{i}}\widetilde{x}_{i}\right\}_{+},
\end{array}
\end{equation}
\end{small}where $\alpha_{j}=max\left(\alpha_{0}, \frac{\sum_{i=1}^{M}\delta(j=y_{i},i\in S_{p\times Intra(j)})\widetilde{W}^{T}_{j}\widetilde{x}_{i}}{1+\sum_{i=1}^{M}\delta(j=y_{i},i\in S_{p\times Intra(j)})}\right),$ $\alpha_{0}$ is an initially given margin, $\delta(\cdot)$ is the indicator function where $\delta(\cdot)=1$ if the condition is satisfied and $\delta(\cdot)=0$ for else, $Intra(j)$ is the number of intra-class cosine similarities in class $j$ and these similarities are sorted in descending order, $p$ is a predefined percentage to control the valid number of similarities in each class. We refer to $S_{p\times Intra(j)}$ as the set including the indices of the first $p\times Intra(j)$ similarities. Analytically, this self-adaptive margin strategy is more suitable for the realistic data distribution, relating the margin to the dynamic feature space and largely alleviating the multifarious human labor of adjusting the margin.
\subsection{NLMC Loss}

Accompanying the cosine similarity constraint in previous parts is the changing norm of features and weight vectors in a mini-batch. As illustrated in Fig.~\ref{fig2}, softmax loss is prone to amplifying the norm. The trade-off between dynamic norm and intra-class cosine similarity constraint seems to be harmful to the final testing accuracy computed by the pair cosine similarities. To better exert the power of this constraint in the training process without sacrificing most of the time on amplifying the norm, we normalize both the features and weight vectors of the last inner-product layer before softmax loss to a same value $s$, which is automatically learned as in \cite{wang2017normface}. In this case, the training process will pay more attention to the intra-class cosine similarity constraint, because all the deeply learned features are distributed on a circle with the same radius in each iteration and the angular between them is an appropriate distance metric. The Normalized Large Margin Cosine (NLMC) loss function is formulated as follows:
\begin{small}
\begin{equation}
\begin{array}{l}
\mathcal{L_{NLMC}}=
-\frac{1}{M}\sum\limits^{M}_{i=1}\log\frac{e^{s^{2}\widetilde{W}_{y_{i}}^{T}\widetilde{x}_{i}}}{\sum_{j=1}^{N}e^{s^{2}\widetilde{W}_{j}^{T}\widetilde{x}_{i}}}+\frac{\lambda}{M}\sum\limits^{M}_{i=1}\left\{\alpha-\widetilde{W}^{T}_{y_{i}}\widetilde{x}_{i}\right\}_{+},
\end{array}
\end{equation}
\end{small}where we substitute $s\widetilde{W}_{j}$ for $W_{j}$ and $s\widetilde{x}_{i}$ for $x_{i}$ in original softmax loss.

\subsection{DLMC Loss}

It seems that the intra-class constraint alone is not enough to obtain discriminative features. Inspired by the form of softmax loss, we extend the NLMC loss to Discriminative Large Margin Cosine (DLMC) loss, which aims to enforce the intra-class cosine similarity larger than the mean of $p\times Inter(j)$ nearest neighboring inter-class cosine similarities with a fixed margin in the normalized exponential feature space.
\begin{small}
\begin{equation}
\begin{array}{l}
\mathcal{L_{DLMC}}=-\frac{1}{M}\sum\limits^{M}_{i=1}\log\frac{e^{s^{2}\widetilde{W}_{y_{i}}^{T}\widetilde{x}_{i}}}{\sum_{j=1}^{N}e^{s^{2}\widetilde{W}_{j}^{T}\widetilde{x}_{i}}}+\frac{\lambda}{M}\sum\limits^{M}_{i=1}\left\{-\log\frac{e^{\widetilde{W}_{y_{i}}^{T}\widetilde{x}_{i}-\alpha}}{\sum_{j=1}^{p\times Inter(y_{i})}e^{\frac{\widetilde{W}_{j}^{T}\widetilde{x}_{i}}{p\times Inter(y_{i})}}}\right\}_{+},
\end{array}
\end{equation}
\end{small}where $Inter(j)$ is the number of different inter-class cosine similarities between a sample of class $j$ and the weight vectors of other classes in a mini-batch, and these cosine similarities are sorted in descending order, $\alpha$ is a predefined margin to discriminate the intra-class and inter-class similarities.

For datasets with many classes, most inter-class similarities are useless. While, the proposed neighborhood sampling strategy can incorporate the most meaningful classes to acquire the reliable mean inter-class similarity. Specifically, when $p\times Inter(j)=1$, the DLMC loss immediately reduces to a variant of triplet loss.
\begin{small}
\begin{equation}
\begin{array}{l}
\mathcal{L_{DLMC}^{'}}=-\frac{1}{M}\sum\limits^{M}_{i=1}\log\frac{e^{s^{2}\widetilde{W}_{y_{i}}^{T}\widetilde{x}_{i}}}{\sum_{j=1}^{N}e^{s^{2}\widetilde{W}_{j}^{T}\widetilde{x}_{i}}}+\frac{\lambda}{M}\sum\limits^{M}_{i=1}\left\{\widetilde{W}_{j}^{T}\widetilde{x}_{i}-\widetilde{W}_{y_{i}}^{T}\widetilde{x}_{i}+\alpha\right\}_{+}.
\end{array}
\end{equation}
\end{small}

Compared to the Euclidean distance constraint in the original feature space of triplet loss, this variant loss function imposes the cosine similarity constraint between a sample and the weight vectors in the normalized feature space, analytically strengthening the robustness in the training process.

\section{Experiments}
The implementation details are given in Section 3.1. In Section 3.2, some exploratory experiments are conducted to find the best settings of hyper-parameters in each method. Finally, we evaluate our approaches on three face recognition benchmark datasets in Section 3.3 and 3.4.

\subsection{Implementation Details}

\textbf{Basic Training Settings.} To test the sensitivity of face recognition results regarding different face detectors, we preprocess the face images by MTCNN \cite{zhang2016joint} and SeetaFace \cite{wu2017funnel} detectors, respectively. We use the publicly available CASIA-WebFace \cite{yi2014learning} as the training set, which originally has 494,414 labeled face images from 10,575 individuals. After removing the undetected images, the resulting datasets have 490,869 images for MTCNN and 437,633 images for SeetaFace. The obvious difference between these two detectors is that there is a high false negative rate of SeetaFace, such that the resulting training set has few false positive samples. We use the Caffe library \cite{jia2014caffe} to implement the CNN model \cite{wen2016discriminative} in this paper, which is a reduced version of ResNet with only 27 convolutional layers. The input faces are cropped to $112\times96$ RGB images, followed by subtracting 127.5 and dividing by 128. The batch size is set to 256 in all the experiments, and the images are horizontally flipped for data augmentation. For LMC, HLMC and MALMC, we train the models from scratch. The initial learning rate is set to 0.1, then divided by 10 at 16K, 24K iterations. The complete training terminates at 28K iterations. While, we fine-tune the networks of NLMC, NLMC+MALMC and DLMC from the softmax baseline model and a relatively small learning rate of 0.001 is applied. For other compared metric loss functions, we train them to achieve their best performance. The classical back-propagation algorithm and mini-batch based Stochastic Gradient Descent (SGD) work well for the training, and the momentum and weight decay are set to 0.9 and 0.0005.

\textbf{Evaluation.} The proposed methods are evaluated on three face recognition datasets, namely LFW, YTF and IJB-A datasets. 10-fold validation is used to acquire the final performance. We extract the features from both the frontal face and its mirror image, and merge the two features by element-wise summation. PCA dimension reduction is applied to the final representations. Nearest neighbor and threshold comparison are used for both identification and verification tasks. Note that we only use single model for all the testings.

\subsection{Exploratory Experiments}
All the experiments in this section are conducted on the resulting dataset by MTCNN detector, if not specified.

\begin{figure}[h]
\centering
\vspace{-0.3cm}
\subfigure[]{
\label{a}
 \includegraphics[height=5cm,width=6.7cm,trim=40 0 0 0,clip]
 {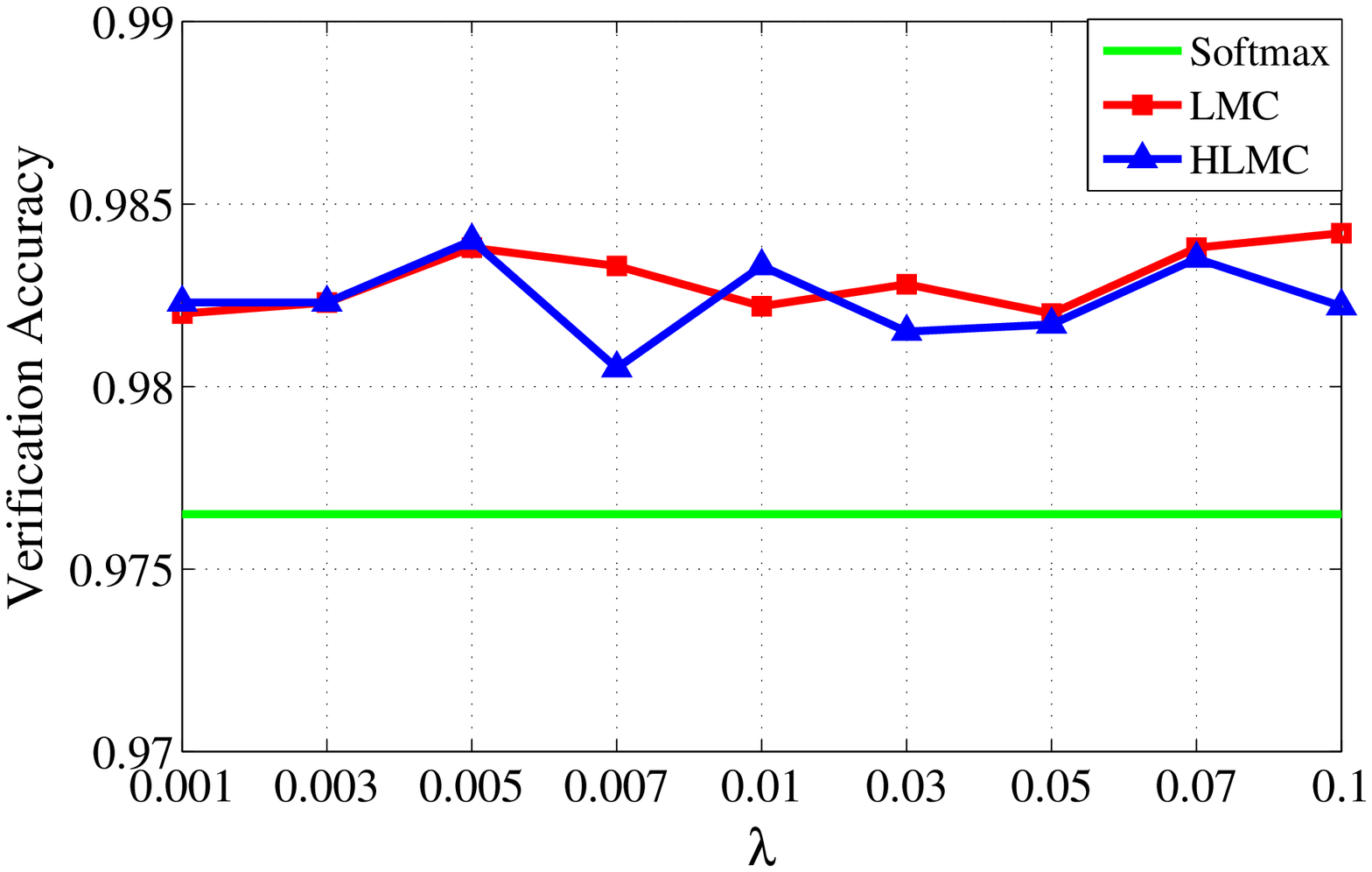}}
 \subfigure[]{
 \label{b}
 \includegraphics[height=5cm,width=6.5cm,trim=40 0 0 0,clip]
 {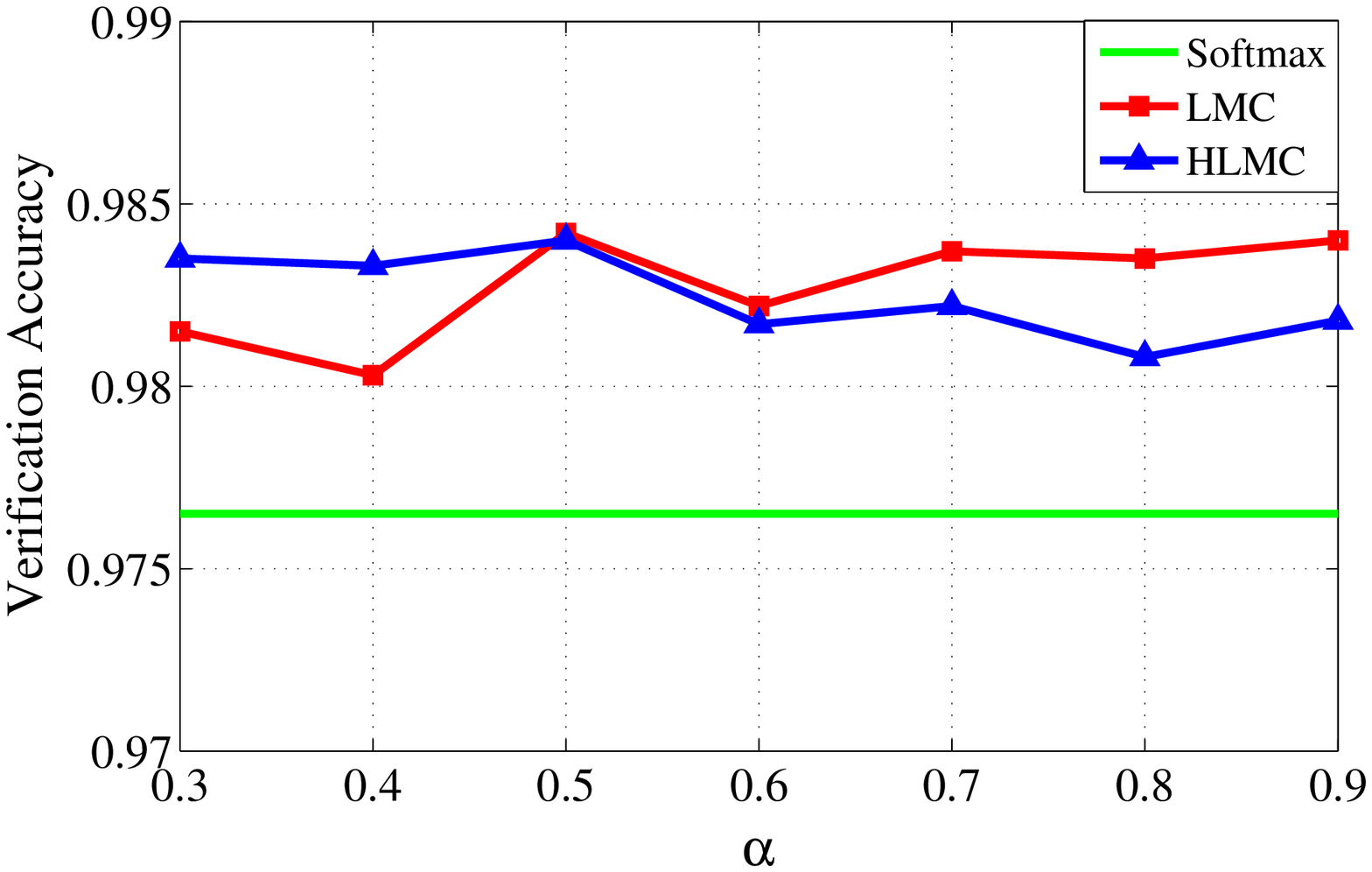}}
 \caption{Verification accuracies on LFW of LMC and HLMC (a) with different $\lambda$ and fixed $\alpha=0.5$. (b) with different $\alpha$, $\lambda=0.1$ for LMC and $\lambda=0.005$ for HLMC.}\label{fig4}
\end{figure}

\textbf{Effect of the hard sample mining strategy.} To ensure the learning efficiency in the training process, we explore a new hard sample mining strategy, where the hard samples refer to the ones misclassified by softmax loss. The hyper-parameters $\lambda$ and $\alpha$ dominate the balance between intra-class and inter-class variations. Properly selected values of them can improve the performance of the proposed methods. So we conduct a pair of contrastive experiments on LMC and HLMC to investigate the sensitivity of these two parameters (Fig.~\ref{fig4}).

In this experiment, we can see that both LMC and HLMC perform much better than the softmax loss. The accuracies fluctuate with different $\lambda$ and $\alpha$. The best settings are $\lambda=0.1, \alpha=0.5$ ($98.42\%$ on LFW) for LMC and $\lambda=0.005, \alpha=0.5$ ($98.40\%$ on LFW) for HLMC. Though the accuracies are almost the same, HLMC simplifies the training process by discarding the easy samples, and we will not deeply explore it in the following experiments.

\begin{figure}[h]
\centering
\vspace{-0.3cm}
\subfigure[]{
\label{a}
 \includegraphics[height=5cm,width=6.5cm,trim=40 0 0 0,clip] 
 {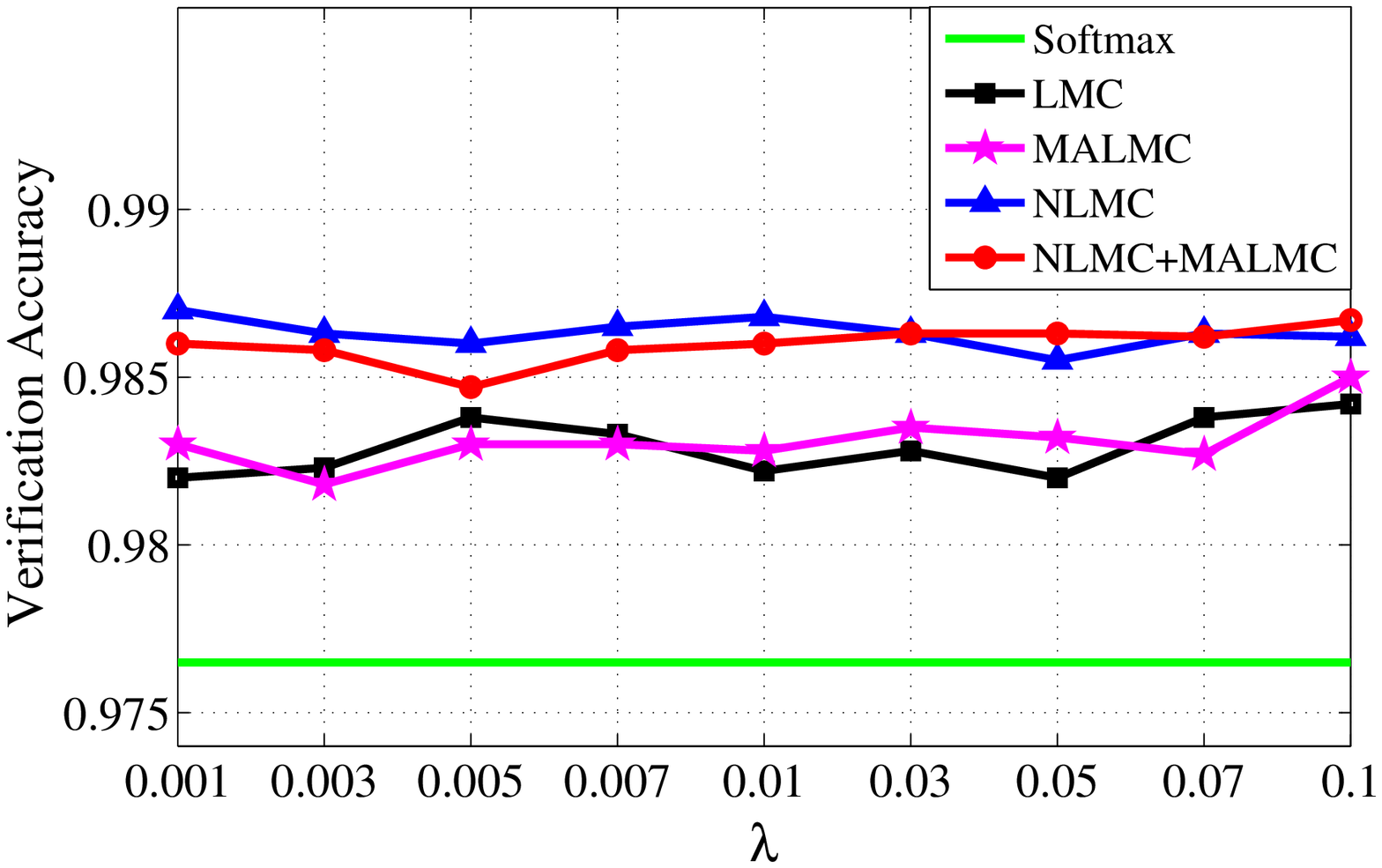}}
 \subfigure[]{
 \label {b}
 \includegraphics[height=5cm,width=6.5cm,trim=35 0 0 0,clip]
 {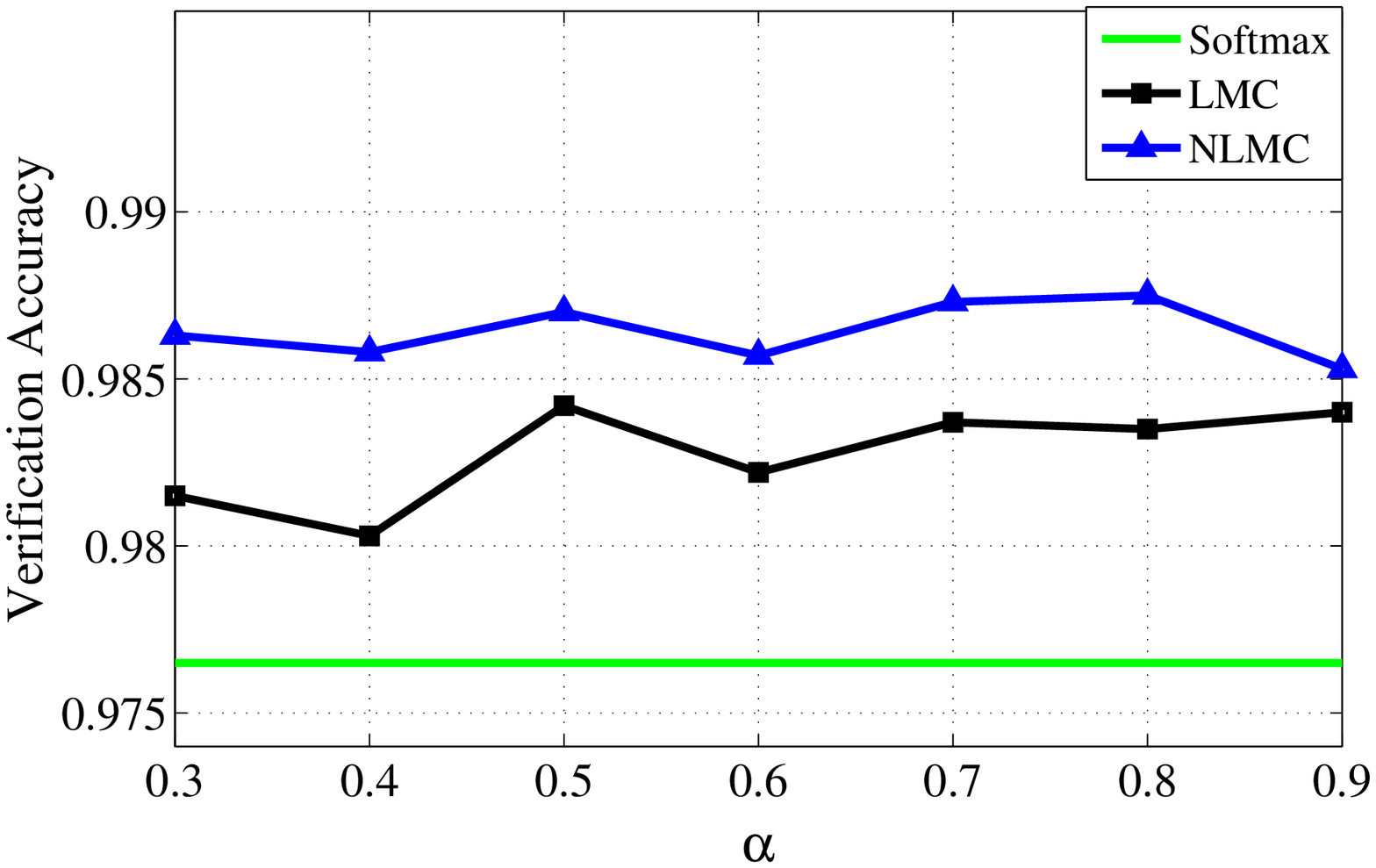}}
 \caption{Verification accuracies on LFW of some proposed methods (a) with different $\lambda$ and fixed $\alpha_{0}=0.2, \alpha=0.5$. (b) with different $\alpha$ and the best setting of $\lambda$ in each method according to (a).}\label{fig5}
\end{figure}

\textbf{Effect of $\lambda$ and $\alpha$.} We can find the importance of choosing the appropriate values of $\lambda$ and $\alpha$ in Fig.~\ref{fig5}. In this section, we explore the best settings of these two hyper-parameters in some of the proposed methods.

In the first experiment, we fix $\alpha$ to $0.5$ and vary $\lambda$ from $0.001$ to $0.1$. In the second experiment, we fix $\lambda$ as their respective best settings in the first experiment (0.005 for LMC and 0.001 for NLMC) and vary $\alpha$ from $0.3$ to $0.9$. Specifically, we set $\alpha_{0}=0.2$ and $p=0.6$ in MALMC. We can observe that the performance of our models is always stable with different $\lambda$ and $\alpha$, and simply using the softmax loss is not a good choice.

\begin{figure}[htp!]
\centering
\vspace{-0.3cm}
\subfigure[]{
\label{a}
 \includegraphics[height=4.5cm,width=6.5cm,trim=35 0 0 0,clip] 
 {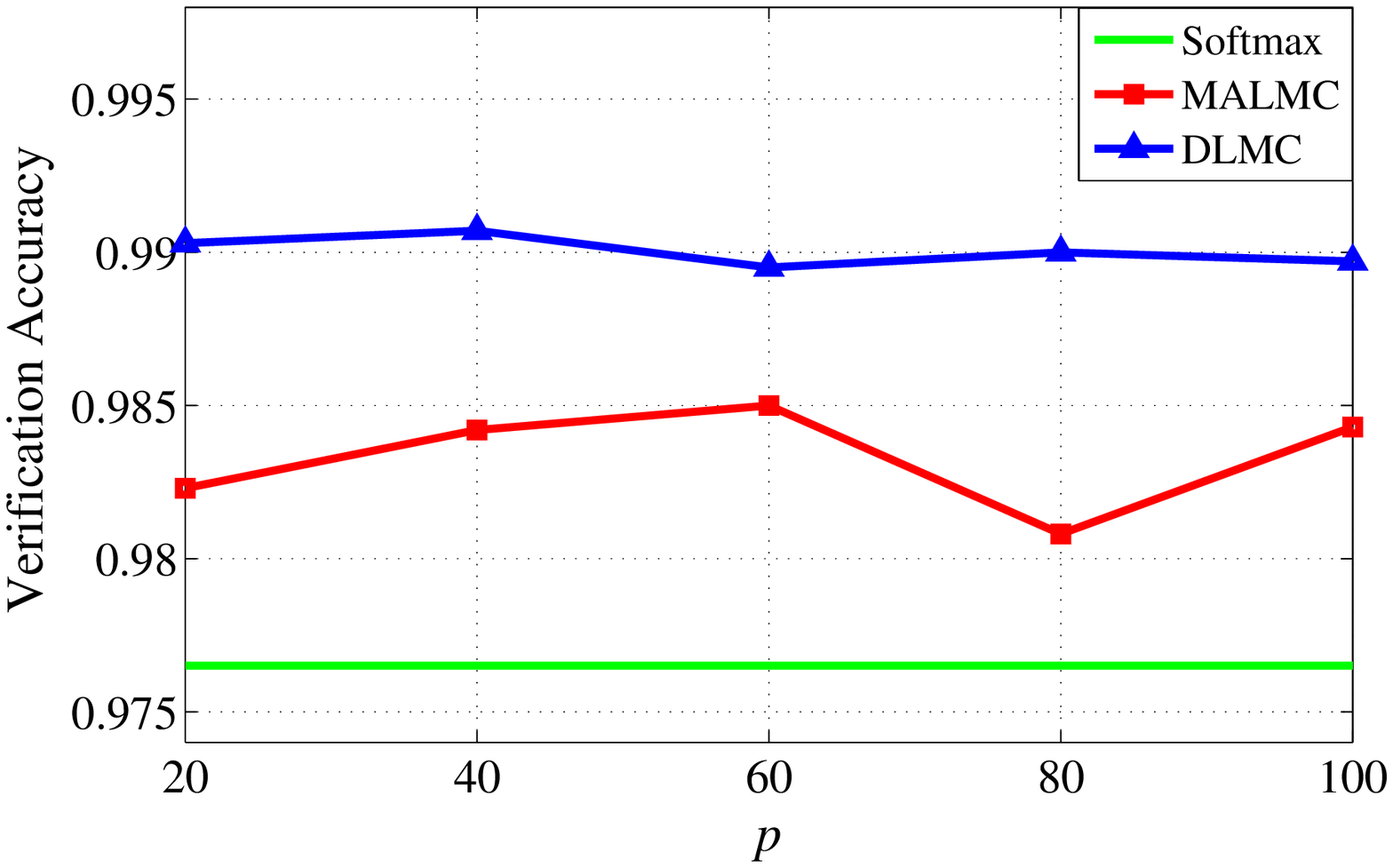}}
 \subfigure[]{
 \label {b}
 \includegraphics[height=4.5cm,width=6.5cm,trim=40 0 0 0,clip]
 {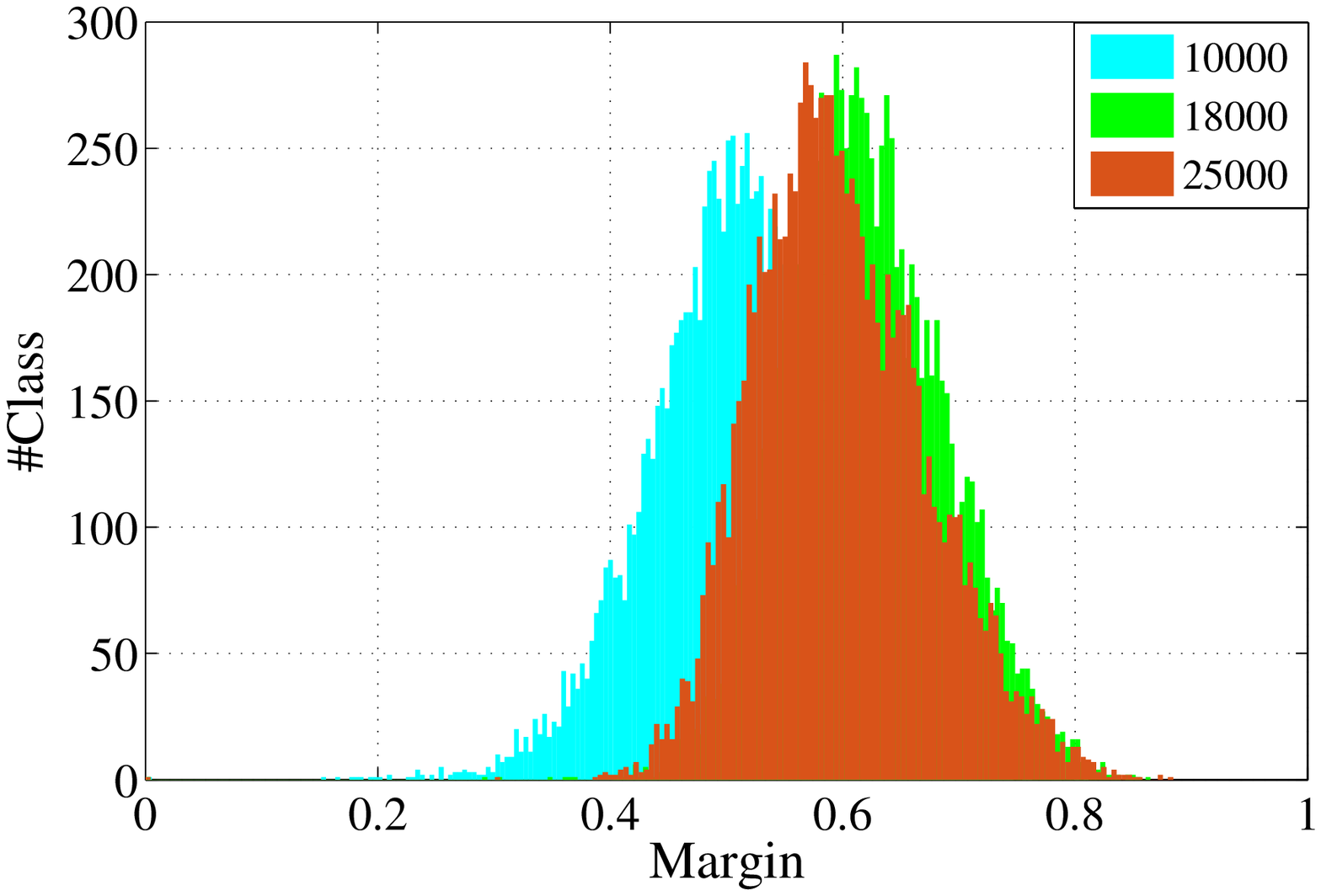}}
 \caption{(a) Verification accuracies on LFW with different $p$, fixed $\lambda=0.1$ for MALMC by MTCNN and $\lambda=0.03,\alpha=0.01$ for DLMC by SeetaFace. (b) The margin distribution of each class with different training iteration steps in MALMC, where $\lambda=0.1, \alpha_{0}=0.2$ and $p=0.6$.}\label{fig6}
\end{figure}

\textbf{Effect of $p$ in MALMC and DLMC.} In this section, we explore the effect of different neighbors on the performance of MALMC and DLMC, namely the verification accuracies on LFW with different $p$ in MALMC and DLMC, while keeping other parameters fixed as their best settings in the previous experiments (Fig.~\ref{fig6}a).

It is obvious that MALMC is sensitive to $p$, and its best setting is 0.6, which controls the valid number of intra-class similarities in a mini-batch. While, DLMC is robust to $p$ across a wide range. The reason is that there exists an inconsistent distribution in each mini-batch, a fixed $p$ dose not seem suitable for measuring the updated feature subspace of each class. However, the robustness of DLMC stems from its similarity to softmax loss which is accompanied by significant inter-class separability, so that the first largest inter-class similarities play the most important role in the training process.

\textbf{Self-adaptive margin strategy in MALMC.} To make clear the margin updating process in MALMC, we perform a toy example of the margin statistics of each class with different iteration steps (10000, 18000 and 25000) during training (Fig.~\ref{fig6}b). One can find that the margin is prone to be larger, and eventually fluctuates around the best value of $0.6$.

\subsection{Experiments on LFW and YTF datasets}

$\mathbf{LFW}\  \   $This dataset contains 13,233 face images of 5,749 different identities from the Internet, with large variations in pose, expression and illumination. For comparison, algorithms typically report the mean face verification accuracies and the ROC curves on 6,000 given face pairs, following the standard protocol of unrestricted with labeled outside data \cite{huang2014labeled}.

$\mathbf{YTF}\  \   $This dataset consists of 3,425 videos from 1,595 different people, with an average of 2.15 videos for each identity. Just as the experiments on LFW, we follow the standard protocol of unrestricted with labeled outside data \cite{wolf2011face}, and report the results on 5,000 video pairs. The final similarity score of each video pair is computed by the average of the cosine similarities from 100 frame pairs.

\begin{table}[h]
\centering
\caption{Face verification performance on LFW and YTF datasets, where [m] refers to the result by MTCNN detector and [s] refers to the result by SeetaFace detector.}\label{table2}
\scalebox{0.75}[0.75]{
\begin{tabular}{ l|c|c|c|c|c }
\hline
\hline
\textbf{Method}   &   \textbf{\#Alig.}    &   \textbf{\#Train}    &  \textbf{\#Net} &   \textbf{Acc.~ on~ LFW} (\%)   &   \textbf{Acc.~ on~ YTF} (\%)\\
\hline
High-dim LBP    &   27  &   100K    &   -   &   $95.17$ &   -\\
$\mathrm{DeepFace}$   &   73  &   4M  &   3   &   97.35   &   91.40\\
$\mathrm{Gaussian~ Face}$   &   -   &   20K &   1   &   98.52   &   -\\
$\mathrm{DeepID}$    &   5   &   200K    &   1   &   97.45   &   -\\

DeepID-2+  &   18  &   300K    &   25  &   99.47   &   93.20\\
Center Loss    &   5   &   700K    &   1   &   99.28   &   94.90\\
$\mathrm{FaceNet}$     &   -   &   200M    &   1   &   $\mathbf{99.63}$    &   $\mathbf{95.10}$\\


CASIA-WebFace &   2   &   WebFace    &   1   &   97.73   &   90.60\\
\hline
$\mathrm{Softmax}$[m]  &   5   &   490K    &   1   &   97.65   & 92.26  \\

$\mathrm{Triplet}$[m]   &   5   &   490K    &   1   &   98.12   &  92.96 \\

L-Softmax[m]&   5   &   490K    &   1   &   \textbf{98.98}   &  93.94 \\


$\mathrm{NormFace}$[m]   &   5   &  490K    &   1   &   98.55   &  94.04 \\

$\mathrm{LMC}$[m]  &   5   &   490K    &   1   &   98.42   &  93.30 \\

$\mathrm{MALMC}$[m]   &   5   &   490K    &   1   &   98.50   &  93.68 \\

$\mathrm{NLMC}$[m] &   5   &  490K    &   1   &   98.75   &  94.06
 \\

NLMC+MALMC[m]   &   5   &   490K    &   1   &   98.67   &  94.14  \\

$\mathrm{DLMC}$[m]   &   5   &   490K    &   1   &   98.80   &  \textbf{94.16}  \\
\hline

$\mathrm{Softmax}$[s]  &   5   &   430K    &   1   &   97.42   &  91.52 \\

$\mathrm{Triplet}$[s]   &   5   &   430K    &   1   &   98.20   &   92.16 \\

L-Softmax[s]&   5   &   430K    &   1   &   98.86   &  94.14 \\


$\mathrm{NormFace}$[s]   &   5   &   430K    &   1   &   98.57   &  93.74 \\

$\mathrm{LMC}$[s]  &   5   &   430K    &   1   &   98.05   &  93.04 \\

$\mathrm{MALMC}$[s]   &   5   &   430K    &   1   &   98.07   &  92.90 \\

$\mathrm{NLMC}$[s] &   5   &   430K    &   1   &   98.88   & 93.60  \\

NLMC+MALMC[s]   &   5   &   430K    &   1   &   98.93   &  93.78  \\

$\mathrm{DLMC}$[s]   &   5   &   430K    &   1   &   \textbf{99.07}   &  \textbf{94.16}  \\
\hline
\hline
\end{tabular}
}
\end{table}

In this experiment, we test the methods presented in Section 3 on datasets preprocessed by two different face detectors, namely MTCNN and SeetaFace. Some state-of-the-art methods (High-dim \cite{chen2013blessing}, DeepFace \cite{taigman2014deepface}, Gaussian~ Face \cite{lu2014surpassing}, DeepID \cite{sun2014deep}, DeepID-2+ \cite{sun2015deeply}, Center Loss \cite{wen2016discriminative}, FaceNet \cite{schroff2015facenet}, CASIA-WebFace \cite{yi2014learning}) are incorporated as a contrast, even though most of their high performance is achieved by huge training data or model ensemble. As can be observed in Table~\ref{table2}, while using single model trained on the publicly available small dataset, our methods are still competitive with other models using high-quality private datasets, such as DeepFace (4M) and FaceNet (200M).
\begin{figure}[h]
 \centering
 \vspace{-0.3cm}
  \subfigure[LFW by MTCNN]{
 \label{a}
 \includegraphics[height=5cm,width=6.5cm,trim=40 0 0 0,clip]
 {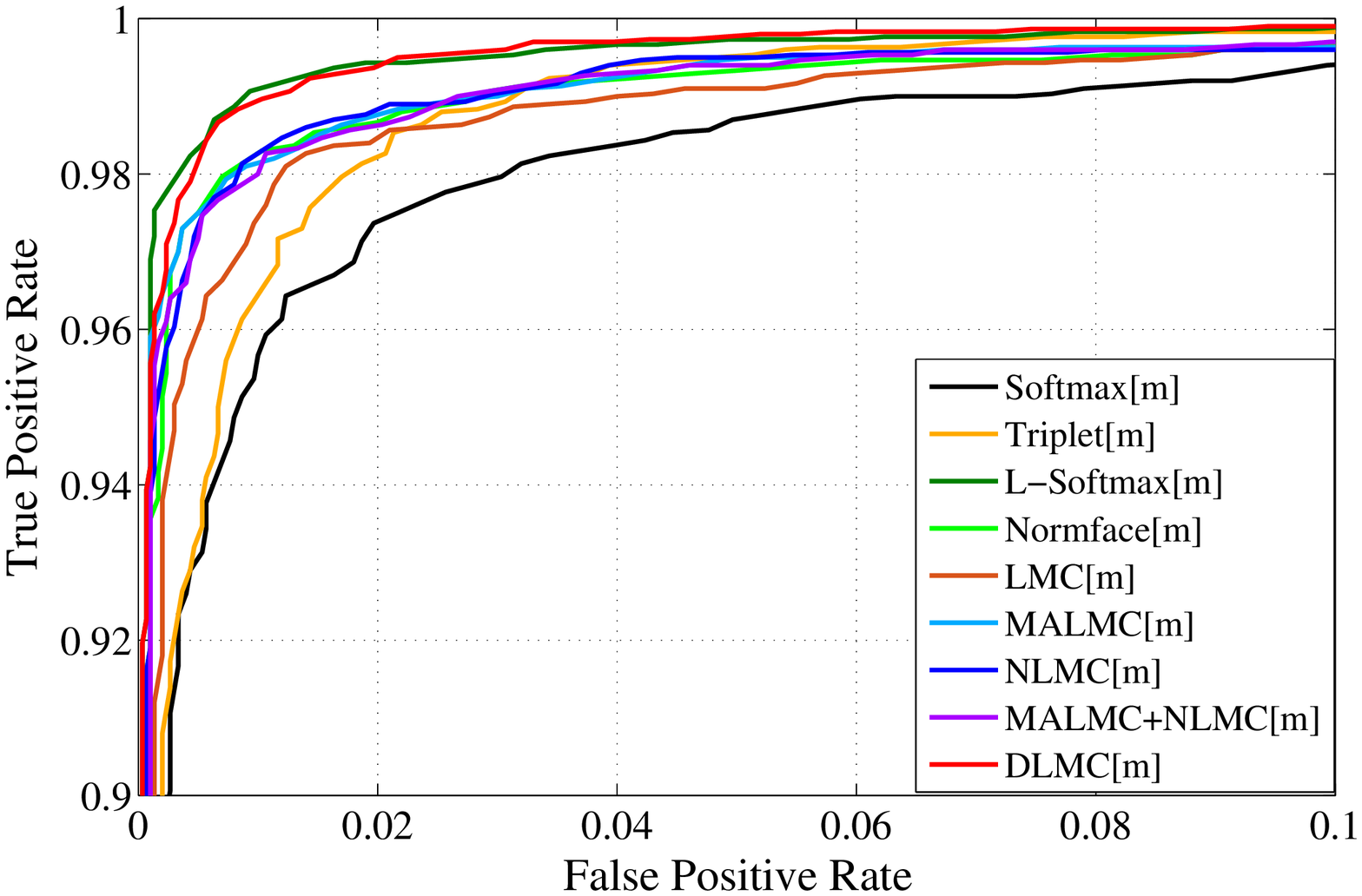}}
 \subfigure[YTF by MTCNN]{
 \label{b}
 \includegraphics[height=5cm,width=6.5cm,trim=40 0 0 0,clip]
 {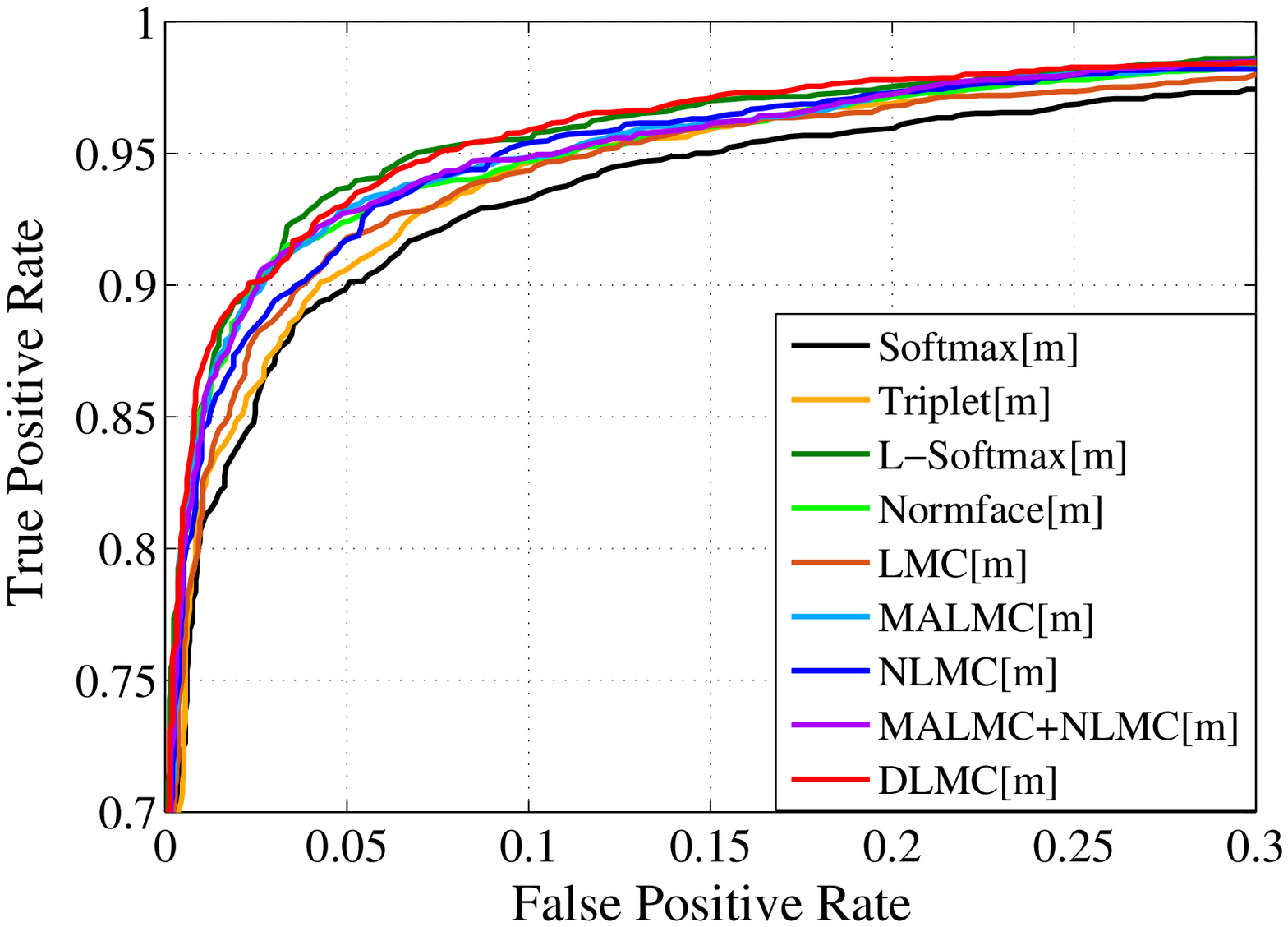}}
  \subfigure[LFW by SeetaFace]{
 \label{c}
 \includegraphics[height=5cm,width=6.5cm,trim=40 0 0 0,clip]
 {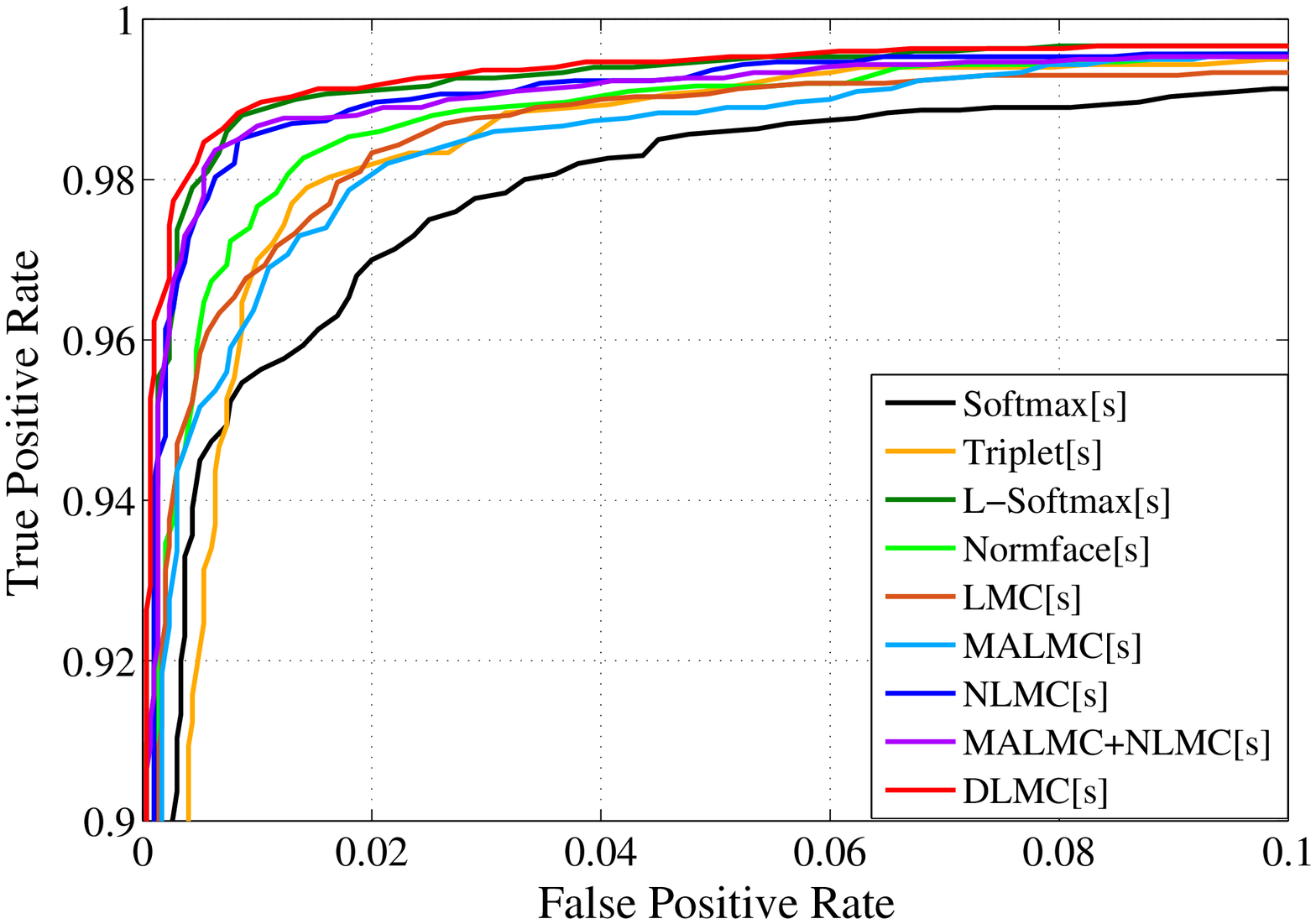}}
 \subfigure[YTF by SeetaFace]{
 \label {d}
 \includegraphics[height=5cm,width=6.5cm,trim=40 0 0 0,clip]
 {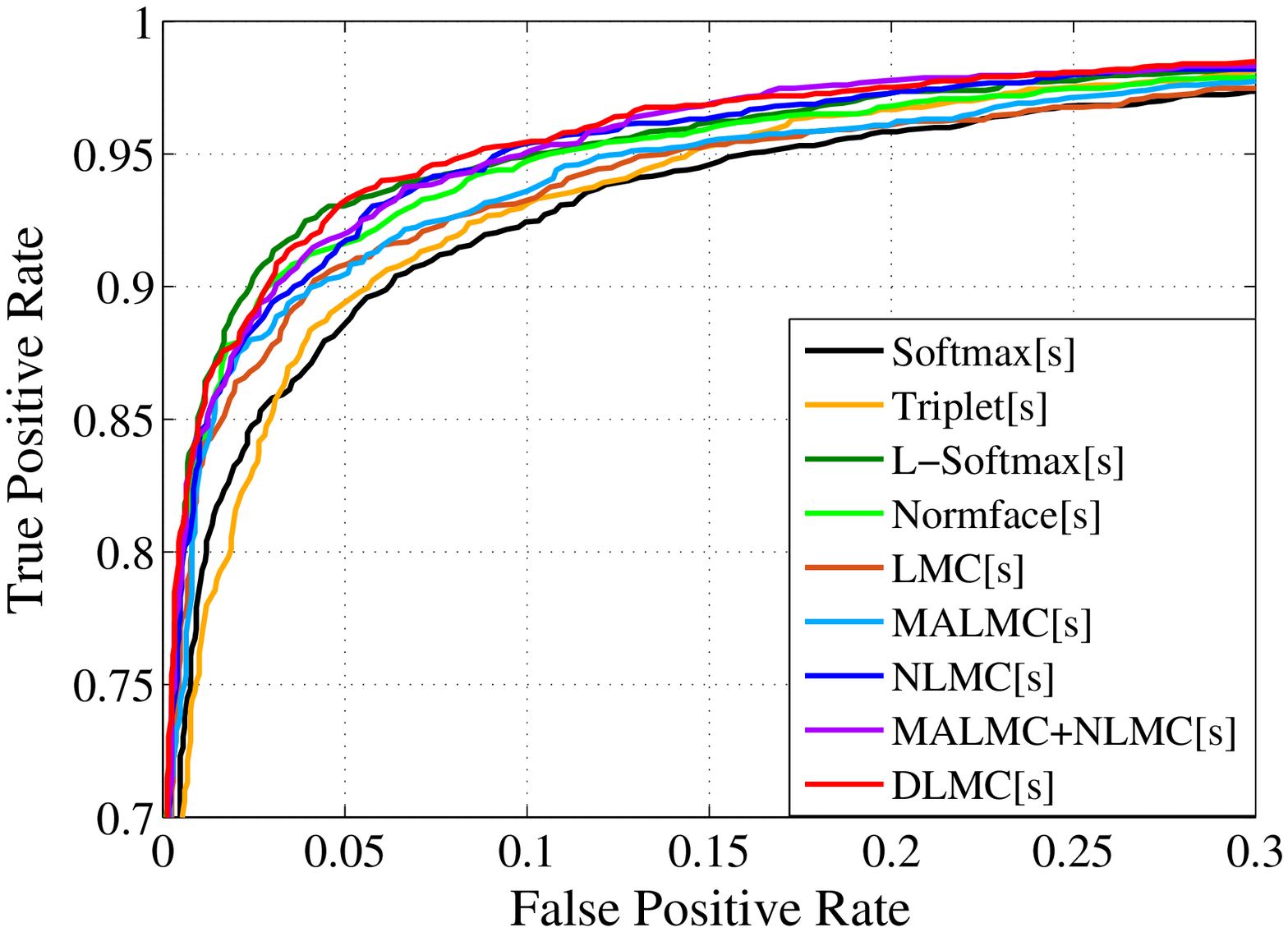}}
 \caption{ROC curves of compared metric loss functions on LFW and YTF datasets by two different face detectors.}\label{fig8}
\end{figure}

For a fair comparison, some typical metric loss functions (Triplet \cite{schroff2015facenet}, L-Softmax \cite{liu2016large}, NormFace \cite{wang2017normface}) are also tested in our own settings. Among these compared loss functions, the proposed methods consistently outperform softmax loss by a significant margin. Specifically, the DLMC loss performs superior by MTCNN ($98.80\%$ accuracy on LFW and $94.16\%$ accuracy on YTF) and SeetaFace ($99.07\%$ accuracy on LFW and $94.16\%$ accuracy on YTF). Compared with NormFace, the NLMC, NLMC+MALMC and DLMC methods obviously show the advantages of cosine similarity constraint in the training process. Similarly, the performance of triplet loss is also not satisfactory. As illustrated in Section 2, the DLMC loss immediately reduces to a variant of triplet loss when $p\times Inter(j)=1$. Besides, the hard triplet mining strategy is avoided here, largely reduces the exponentially increased computational complexity of training dataset. The results convincingly demonstrate that the DLMC loss can alleviates the difficult convergence and big data dependence of triplet loss. The Receiver Operating Characteristic (ROC) curves of them are shown in Fig.~\ref{fig8}. One should notice that there exists a discrepancy between the results of the two different face detectors, and the trends vary from one loss to another. Though, our DLMC method always among the top performance.

\subsection{Experiments on IJB-A dataset}

$\mathbf{IJB}$-$\mathbf{A}\  \   $This dataset contains 5,712 images and 2,085 videos of 500 subjects, with an average of 11.4 images and 4.2 videos per subject. The IJB-A evaluation protocol consists of open-set verification (1:1 comparison) and identification (1:N search) over 10 random training and testing splits. Unlike the LFW and YTF datasets, the IJB-A dataset divides the testing images/video frames into gallery and probe sets, and the subjects are described by templates. Moreover, the images in the IJB-A dataset contain extreme pose, illumination and expression variations without being filtered by a commercial face detector. These factors essentially make IJB-A a challenging unconstrained face recognition dataset \cite{klare2015pushing}. We use the Softmax operator \cite{masi2016we} to compute the similarity score of two sets described by templates.
\begin{table}[h]
\centering
\tabcolsep 12pt
\caption{Results on IJB-A dataset. The True Accept Rate (TAR) at False Accept Rate (FAR)=0.01 and 0.001 for the ROC curves. The Rank-1, Rank-5 and Rank-10 retrieval accuracies for the CMC curves.}\label{table4}
\scalebox{0.75}[0.75]{
\begin{tabular}{l||c|c||c|c|c}
\hline
\hline
\multirow{2}{*}{\textbf{Method}}   &   \multicolumn{2}{c||}{\textbf{IJB-A Ver.(TAR)} (\%)}    &   \multicolumn{3}{c}{\textbf{IJB-A Id.(Rec. Rate)} (\%)}  \\
\cline{2-6}
  &   FAR 0.01 &   FAR 0.001    &   Rank-1  &   Rank-5  &   Rank-10\\
\hline
$\mathrm{GOTS}$ &   40.6   &   19.8   &   44.3   &  59.5 &  -  \\

Deep Multi-Pose &   87.6   &   -   &   84.6   &  \textbf{92.7}   &  94.7  \\

$\mathrm{Template~Adaptation}$ &   \textbf{93.9}   &   \textbf{83.6}   &   92.8   &   -   &    98.6 \\

All-In-One Face &   92.2   &   82.3   &   \textbf{94.7}   & -  &  \textbf{98.8}  \\
\hline
$\mathrm{Softmax}$[m]  &  81.36    &   59.46   & 87.70  & 94.29  & 96.25 \\

$\mathrm{Triplet}$[m] &   70.66   &   47.48   &  88.82  & 94.40  & 96.06 \\

L-Softmax~[m]&   72.71   &   40.10   & 88.45  & 93.85  & 95.71 \\


$\mathrm{NormFace}$ [m]   &   85.86   &   69.77   & 90.98  & 95.51  &  96.76 \\

$\mathrm{LMC[m]}$  &  \textbf{86.47}    &   \textbf{72.71}   & 89.66  & 95.02  & 96.73 \\

$\mathrm{MALMC[m]}$   &   85.41   &   67.55   &  91.15 &  95.77 &  96.90 \\

$\mathrm{NLMC[m]}$ &   86.19   &   70.94   & 90.96  & 95.45  &  96.72  \\

NLMC+MALMC[m]   &   85.52   &   70.05   &  90.63  &  95.41 & 96.73  \\
$\mathrm{DLMC[m]}$   &   86.02   &   62.45   &  \textbf{93.21}  & \textbf{97.34}  & \textbf{98.33}  \\
\hline
\hline
\end{tabular}
}
\end{table}

\begin{figure}[htp!]
 \centering
 \subfigure[]{
 \label{a}
 \includegraphics[height=5cm,width=6.5cm,trim=45 0 0 0,clip]
 {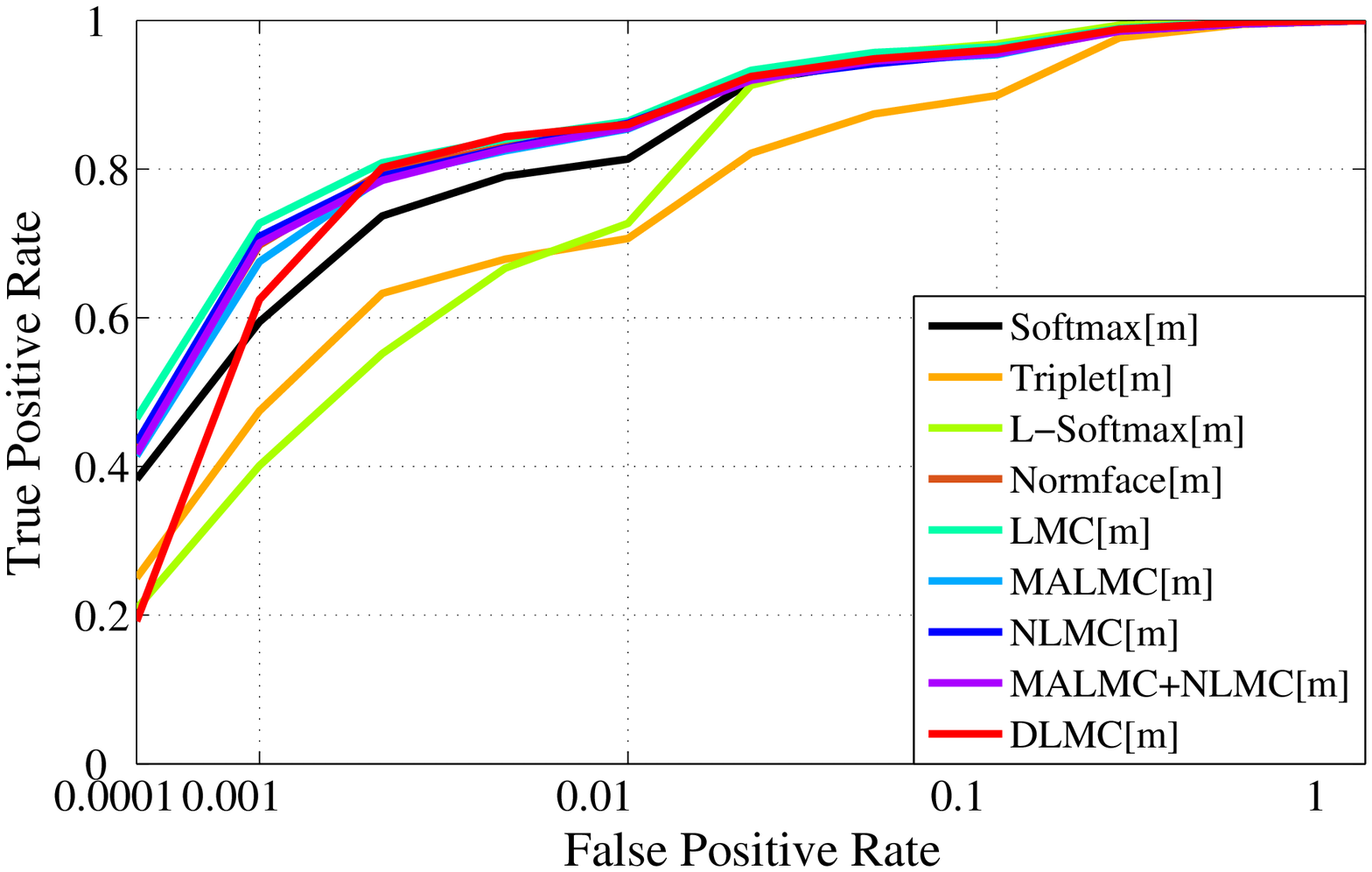}}
 \subfigure[]{
 \label{b}
 \includegraphics[height=5cm,width=6.5cm,trim=45 0 0 0,clip]
 {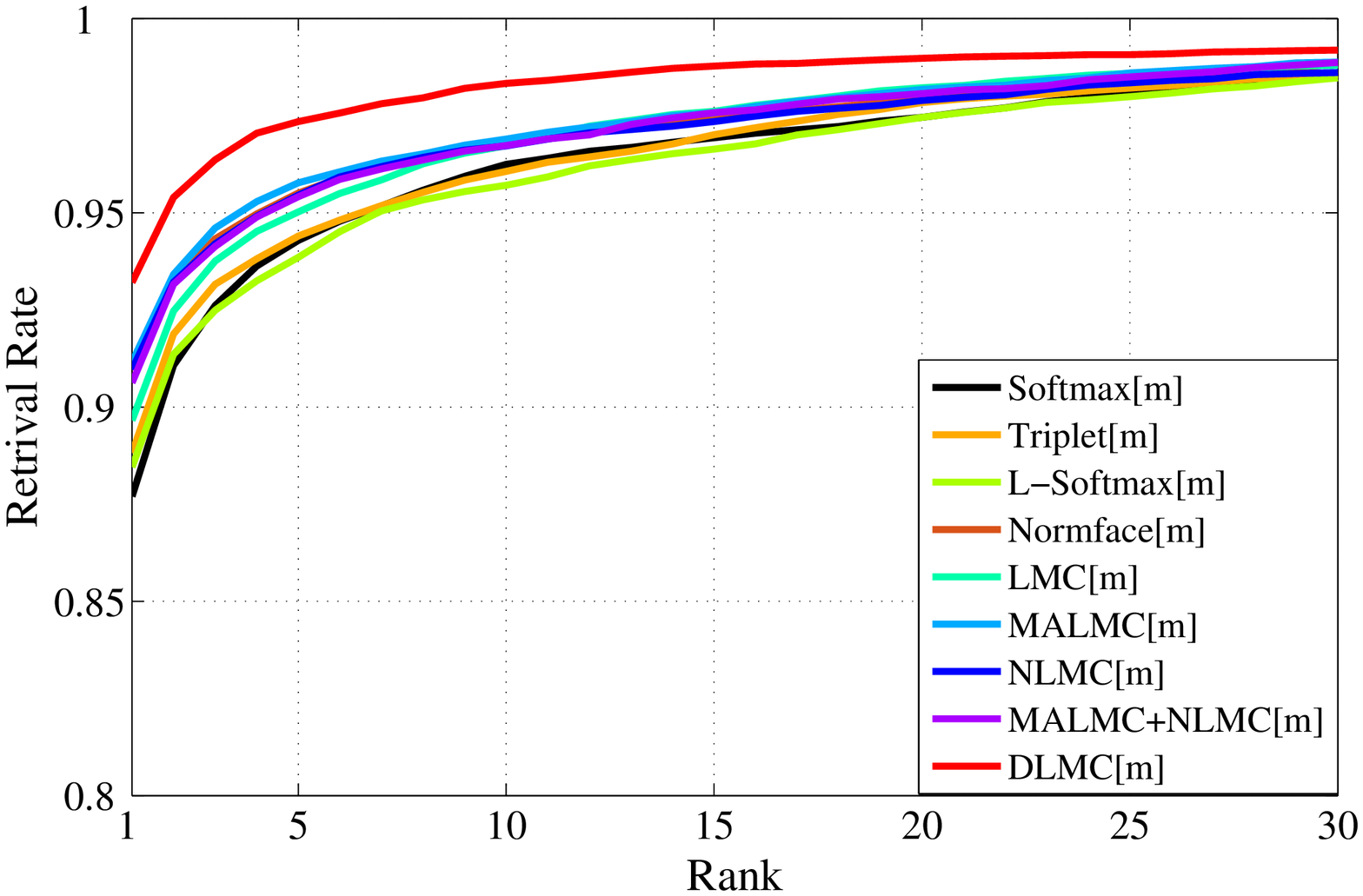}}
 \caption{Recognition accuracies on IJB-A dataset. (a) ROC curves for the compare protocol. (b) CMC curves for the search protocol.}\label{fig9}
\end{figure}

For simplicity, we only present the results by MTCNN detector here. As in the experiments of LFW and YTF, we compare our methods with several current mainstream DML approaches (Triplet, L-Softmax, NormFace) under the same settings, and other state-of-the-art approaches (GOTS \cite{klare2015pushing}, Deep Multi-Pose \cite{abdalmageed2016face}, Template~Adaptation \cite{crosswhite2017template}, All-In-One Face \cite{ranjan2017all}) using larger training datasets or model ensemble. Whereas, simply comparing our methods to those state-of-the-art results is unfair, because their system designs and implementation details are different from ours, and the difficult access to their codes and data makes it hard to say exactly how much improvement our proposed methods acquire. From the results in Table~\ref{table4}, we can find that our proposed methods significantly improve over the off the shelf commercial systems GOTS. Compared to some deep learning based methods, our approaches still achieve satisfactory performance. To better show the comparison results with some typical DML methods under our own settings, the ROC curves for face verification and the Cumulative Match Characteristic (CMC) curves for face identification are plotted in Fig.~\ref{fig9}, respectively. Obviously, our methods exhibit prominent advantages consistently over other DML methods, and always among the top performance. However, the performance of triplet loss and L-Softmax loss on the IJB-A dataset is not as good as that on the LFW and YTF datasets, due to the large variations of IJB-A.

\section{Conclusion and future work}

In this paper, we introduce the intra-class cosine similarity constraint into the training process, to alleviate the large intra-class variations of softmax loss and keep consistency with the testing process. Accompanied by the inter-class separability of softmax method, the original LMC method achieves a significant improvement. Based on this, the MALMC method is proposed to mitigate the fussy human labor of adjusting the margin hyper-parameter. Furthermore, the NLMC method is given to take full advantage of the intra-class cosine similarity constraint with all the features and weight vectors in a mini-batch fixed to the same norm. To acquire more discriminative features, a profound idea of considering the intra-class and inter-class constraints simultaneously is proposed to form the DLMC method. Extensive experiments on several public face recognition benchmark datasets convincingly demonstrate the effectiveness and robustness of these proposed methods, even on a small training dataset.

Noticeably, these loss functions are not differentiable everywhere, and some smoothed versions seem to be a meaningful research direction. We will apply the proposed methods on other metric leaning tasks in the future, such as person re-identification or image retrieval. Furthermore, how to develop robust DML methods regarding different face detectors is an interesting future direction for research.

\section*{References}
\bibliographystyle{ieeetr}
\bibliography{mybibfile}

\end{document}